\documentclass{article}
\usepackage{spconf,amsmath,graphicx}

\usepackage[dvipsnames]{xcolor}

\title{Reinforcement Learning of Graph Neural Networks for Service Function Chaining}

\name{DongNyeong Heo$^{\star}$ \qquad Doyoung Lee$^{\dagger}$ \qquad Hee-Gon Kim$^{\dagger}$ \qquad Suhyun Park$^{\dagger}$ \qquad Heeyoul Choi$^{\star}$ \thanks{This work was supported by Institute for Information and communications Technology Promotion (IITP) grant funded by the Korea government (MSIT) (No.2018-0-00749, Development of virtual network management technology based on artificial intelligence).}}

\address{$^{\star}$ 
Dept. of Information and Communication Engineering\\
Handong Global University, Pohang, South Korea\\
$^{\dagger}$
Dept. of Computer Science and Engineering \\
Pohang University of Science and Technology, Pohang, South Korea }

\begin{document}

\maketitle
\begin{abstract}
In the management of computer network systems, the service function chaining (SFC) modules play an important role by generating efficient paths for network traffic through physical servers with virtualized network functions (VNF). To provide the highest quality of services, the SFC module should generate a valid path quickly even in various network topology situations including dynamic VNF resources, various requests, and changes of topologies. The previous supervised learning method demonstrated that the network features can be represented by graph neural networks (GNNs) for the SFC task. However, the performance was limited to only the fixed topology with labeled data. In this paper, we apply reinforcement learning methods for training models on various network topologies with unlabeled data.
In the experiments, compared to the previous supervised learning method, the proposed methods demonstrated remarkable flexibility in new topologies without re-designing and re-training, while preserving a similar level of performance.
\end{abstract}


\section{Introduction}
\label{sec:intro}
Development of computer network system based on software such as software defined networking (SDN) and network function virtualization (NFV) has provided many advantages to managing network systems \cite{kreutz:sdn-survey}\cite{mijumbi:nfv-survey}. Due to the advances in such developments, the automatic network controlling system has become a promising technology \cite{datta:intelligent-networking}, which includes some modules such as virtualized network function (VNF) deployment, resource demand prediction, auto-scaling, anomaly detection system, and service function chaining (SFC). Among them, the SFC module carries out the essential task of generating a path that connects the requested VNF instances \cite{Bhamare:sfc-survey}. A path is generated according to active user requests that consist of a chain of VNFs and locations of source and destination. The resulting paths should satisfy not only these requested requirements, but also provide services with the shortest possible delay. In addition, the SFC module should be flexible with various VNF resources statuses, various types of user requests, and various network topologies. Thus, it is a challenging task to develop an efficient SFC module.

Recently, deep learning techniques have been applied to the complicated SFC task with high capacity models. To the best of our knowledge, the first deep learning application to the SFC task was based on simple deep neural networks (DNNs) as introduced in \cite{pei:deep-sfc}, where the performance was not sufficient to use as an actual SFC module. Another deep learning approach to the SFC task was based on the gated graph neural network (GG-NN) \cite{li:ggnn} as in \cite{Heo:supervised_gnn_sfc}. As variants of the graph neural networks (GNNs) \cite{scarselli:gnn}, the GG-NN models could capture the features of the network topology statuses and map them to the node level vector representations based on the topology \cite{Heo:supervised_gnn_sfc}. The GG-NN models outperformed the DNN models in terms of time delay for service and failure ratio for finding paths. Also, the GG-NN models could be applied to even dynamically changing network topologies without re-designing or re-training, since GNNs could accept the variable number of nodes as the input to the networks. 

However, since the GG-NN models in \cite{Heo:supervised_gnn_sfc} were trained with a supervised learning (SL) algorithm, the models needed a large amount of labeled data for training. Actually, it is possible to produce labeled data (true path) from various topologies only if a network topology simulator and the integer linear programming (ILP) algorithm are available. Even in such cases, the labels are not available from real-time network systems because it takes too much time to generate labels. Therefore, the SL approach is inevitably biased to specific topology with labeled data, so that the performance seriously deteriorates in tests on new computer network topology, because the model can not be trained on various topologies.

To make the GG-NN models flexible with various topologies while maintaining the good performance, we propose the use of reinforcement learning (RL) algorithms to train the models with the same architecture as \cite{Heo:supervised_gnn_sfc}. The RL algorithms have the advantage of the agent that can learn from rewards that are provided by the network environment, so that the models can learn from various topologies even when labels are not available. We design a simple reward setting and train the models with the REINFORCE algorithm \cite{Williams:REINFORCE, Ranzato2015}. 
We train the models with randomly changed topologies and randomly placed VNF instances during the training episodes. 

In the experiment with the fixed topology test, the proposed RL approach had a similar level of performance to the SL approach. However, it significantly outperformed the SL approach in tests with changed topologies and changed positions of VNF instances.

\section{Background}
\label{sec:background}
\subsection{Service Function Chaining}
We follow the same problem definition as in \cite{Heo:supervised_gnn_sfc} for the SFC task, and briefly describe it in this section. On the network topology, several VNF instances are deployed on the computing nodes as described in \cite{lange2019predicting}, which deploys a sufficient number of requested VNF resources while minimizing leftovers. According to user requests for a service, the SFC module should generate a resulting path to provide high quality of services (QoS). The QoS is generally measured by packet loss, bit rate, throughput, transmission delay and availability. However, because our experiments are conducted on simulated situations, the QoS is measured by the success of generation and the total time delay. The generated path is classified as a success case if it processes all of the requested VNFs and arrives at the destination node within a pre-defined number of maximum steps. Delays are caused by forwarding network traffic and processing VNFs. Below is the formulation of the total delay time of a path $p_i$.
\begin{equation}
    Delay(p_i) = \sum_{uv \in E} d_{uv}y_{uv}^{p_i} + \sum_{m \in M} d_m x_{m}^{p_i}, \label{eq:delay}
\end{equation}
where $E$ is the set of edges and $M$ is the set of VNF instances in the network topology. $d_{uv}$ and $d_m$ are delay time of that edge and VNF instance, respectively. $y_{uv}^{p_i}$ and $x_m^{p_i}$ indicate that the edge $uv$ and VNF instance $m$ are included in the resulting path $p_i$.

\begin{figure}[t]
    \hbox{\centering \hspace{0.35in}
    \includegraphics[width=0.6\linewidth]{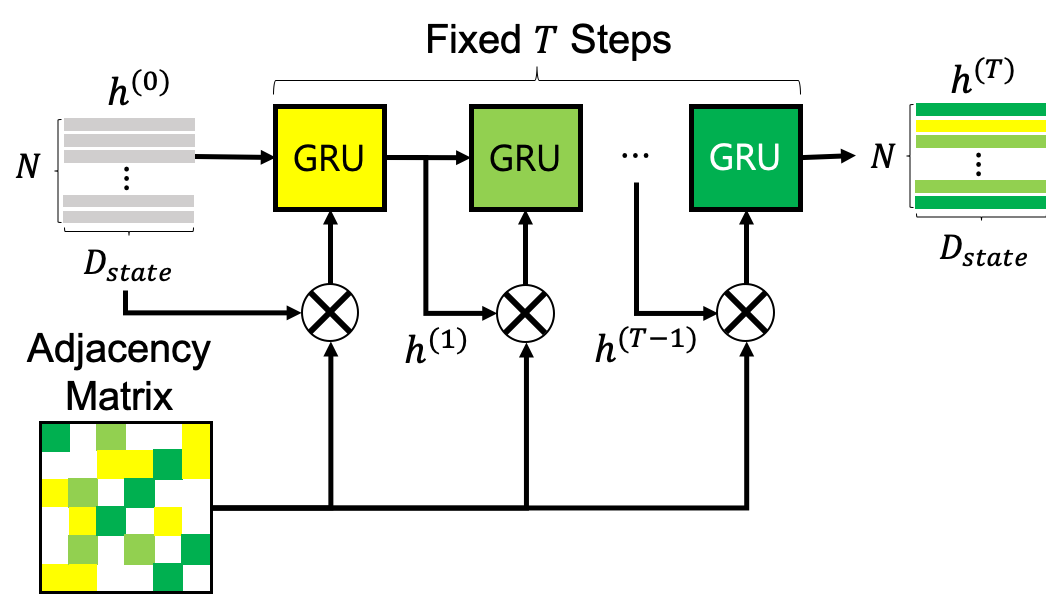}
    }
    \hbox{\small \hspace{1.0in}(a) Encoder Model}
    \vspace{0.1in}
    \hbox{\centering 
    \includegraphics[width=0.9\linewidth]{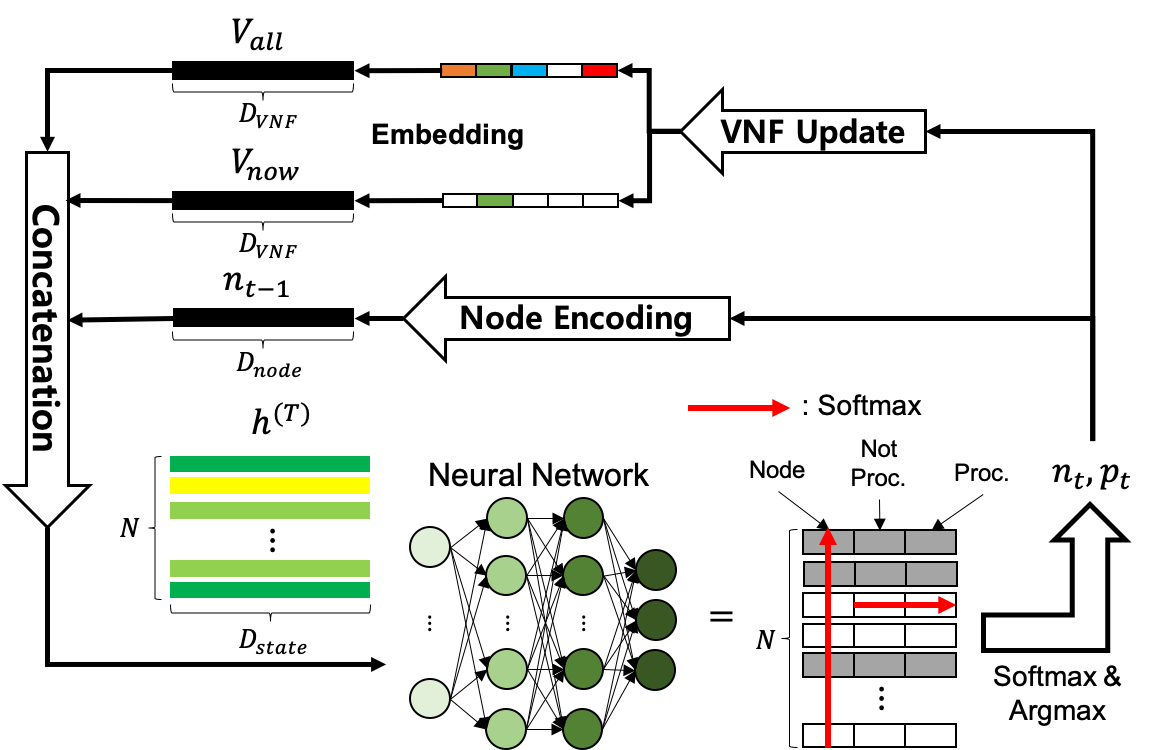}
    }
    \hbox{\small \hspace{1.0in}(b) Decoder Model}
    \caption{GG-RNN model architecture with encoder and decoder. Adapted from \cite{Heo:supervised_gnn_sfc}.}
    \label{fig:model_architecture}
\end{figure}

\subsection{GG-RNN}
\label{subsec:GG-RNN_architecture}
In the GNN-based SFC task \cite{Heo:supervised_gnn_sfc}, they designed a model with the encoder-decoder architecture. Since they used GG-NN \cite{li:ggnn} for the encoder and recurrent neural network (RNN) for the decoder, they referred it to as GG-RNN. In this section, we briefly describe the GG-RNN model as shown in Fig. \ref{fig:model_architecture}. For more details of the model architecture, see \cite{Heo:supervised_gnn_sfc}.

\subsubsection{Encoder}
\label{subsubsec:Encoder}
As an encoder, GG-NN produces a vector representation of nodes as follows:
\begin{align}
    h_u^{(0)}&=[l_u^\top,0]^\top, \label{eq:init_h} \\
    h_{u,in}^{(t)}&=A_u^\top[h_1^{(t-1)\top} \dots h_{|N|}^{(t-1)\top}]^\top, \label{eq:init_adj} \\
    h_u^{(t)}&=GRU(h_u^{(t-1)}, h_{u,in}^{(t)}),  
\end{align}
where $h_u^{(0)}$ is the initial hidden state of node $u$ which is annotation consisting of the label of that node $l_u$, the feature of that node $u$. In Eq. \eqref{eq:init_adj}, $A$ is the adjacency matrix describing the connection between nodes, where $N$ is the number of nodes in the network topology. With the adjacency matrix $A$, Eq. \eqref{eq:init_adj} calculates the information exchange between neighbor nodes so that the output representation of the encoder reflects how the computing nodes and the switches are connected in the network.
The $GRU$ function is the gated recurrent unit (GRU) \cite{cho:gru} with the previous hidden state $h_u^{(t-1)}$ and the current input $h_{u,in}^{(t)}$. 

\subsubsection{Decoder}
\label{subsubsec:Decoder}
Given the final representation of the GG-NN encoder, the decoder receives additional inputs from the network topology. The first one is $V_{all}$ which includes the list of requested VNF types, and the second one is $V_{now}$ for the VNF type with the top priority to process at the current time. The decoder estimates the probabilities of the nodes to forward the network traffic. In addition, it also estimates the probabilities of whether or not to process the top priority VNF instance on the next node. By repeating this step with RNN, the decoder generates a valid path or fails. 

\section{Training of GG-RNN}
\label{sec:proposed_approach}
\subsection{Reinforcement Learning Components}
To apply RL methods for the SFC task, we define the core components of RL: state, action, and reward. The state is composed of the input into the GG-RNN model including annotation, adjacency matrix, and three additional inputs of the decoder as mentioned above. In addition, the previous hidden state of the decoder RNN is also part of the state. The action consists of the selection of the next node and decision to process the VNF at the selected node by the policy network (GG-RNN). We assign a positive reward to the agent only when it succeeds in generating a valid path. For the model to generate a shorter path, we also add a negative total delay time of the path to the reward as a penalty, where $\lambda$ can control the weight of the penalty. The reward can be defined by 
\begin{equation*}
    r_t = \begin{cases}
    10000 - \lambda * Delay(P) &\text{if $t$ is terminal},\\
    0 &\text{otherwise},
    \end{cases}
\end{equation*}
where $P$ is a complete episode with the sequence of transitions $\{s_1, a_1, r_2, \dots s_{T-1}, a_{T-1}, r_T\}$. The total delay is computed by Eq. \eqref{eq:delay}. 

\subsection{REINFORCE algorithm}
\label{subsec:REINFORCE_algorithm}
The REINFORCE algorithm \cite{Williams:REINFORCE} is one of the policy gradient algorithms in reinforcement learning. The agent makes an action on a given state according to its policy. After repeating such transitions with the policy until the state reaches a terminal state, an episode is finished. Then, the update algorithm calculates the returns for all of the actions that are included in the episode from the last transition to the first transition. After these calculations are made, the parameters of the policy network are updated based on the derivative of the log probability of those actions, where the derivative is weighted by the return of the actions. As a result, the policy network is trained to maximize returns for each episode. 

Since the GG-RNN model repeatedly makes an action to find the next node until it completes a path to serve a user request, REINFORCE can be applied to train the model. We can update and compute the returns of the GG-RNN model for a resulting path as follows. 
\begin{align}
    \theta_{new} &= \theta_{old} + \alpha \nabla_\theta \log \text{GG-RNN}_\theta(s_t,a_t)G_t, \label{eq:policynet_update}\\
    G_t &= r_t + \gamma G_{t+1}, \label{eq:returns}
\end{align}
where we use $\text{GG-RNN}_{\theta}(s_t,a_t)$ as a policy network in REINFORCE. In the episode, $s_t, a_t$, and $ r_t$ are state, action and reward at $t$, respectively, as described in the previous section. $G_t$ is the return value at $t$, and $\alpha$ and $\gamma$ are a learning rate and a discount factor, respectively.

\begin{figure}[t]
    \hbox{\centering \hspace{0.7in}
    \includegraphics[width=0.51\linewidth]{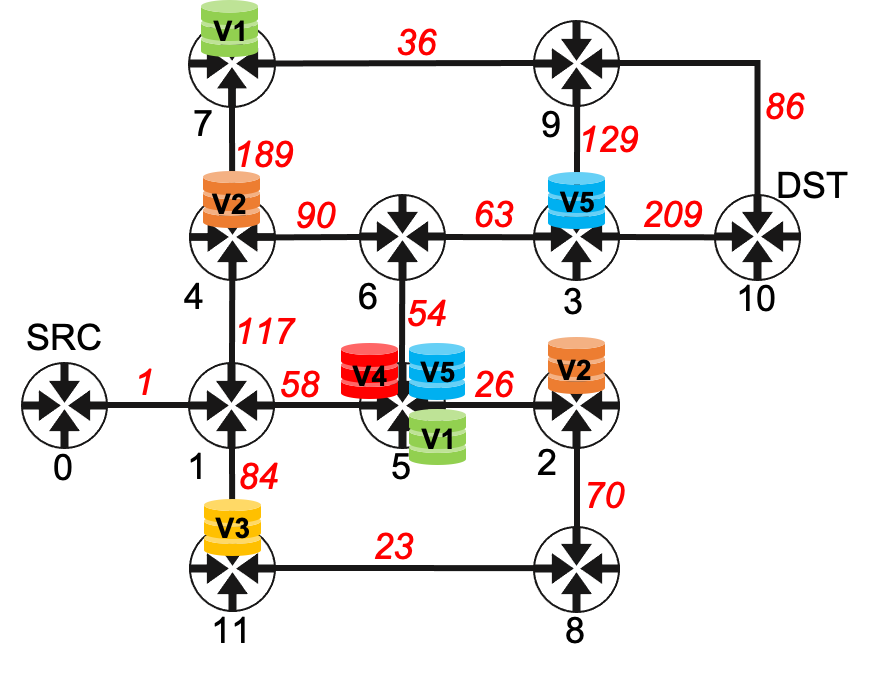}
    }
    \hbox{\small \hspace{1.0in}(a) Original Topology}
    \vspace{0.1in}
    \hbox{\centering \hspace{-0.075in}
    \includegraphics[width=0.51\linewidth]{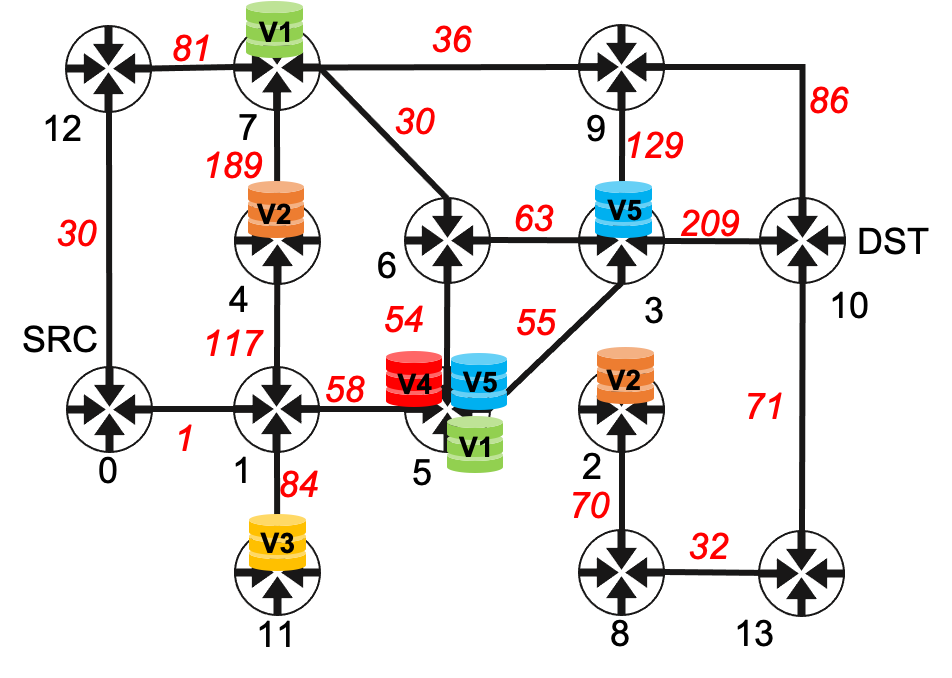}
    \hspace{0.0in}
    \includegraphics[width=0.51\linewidth]{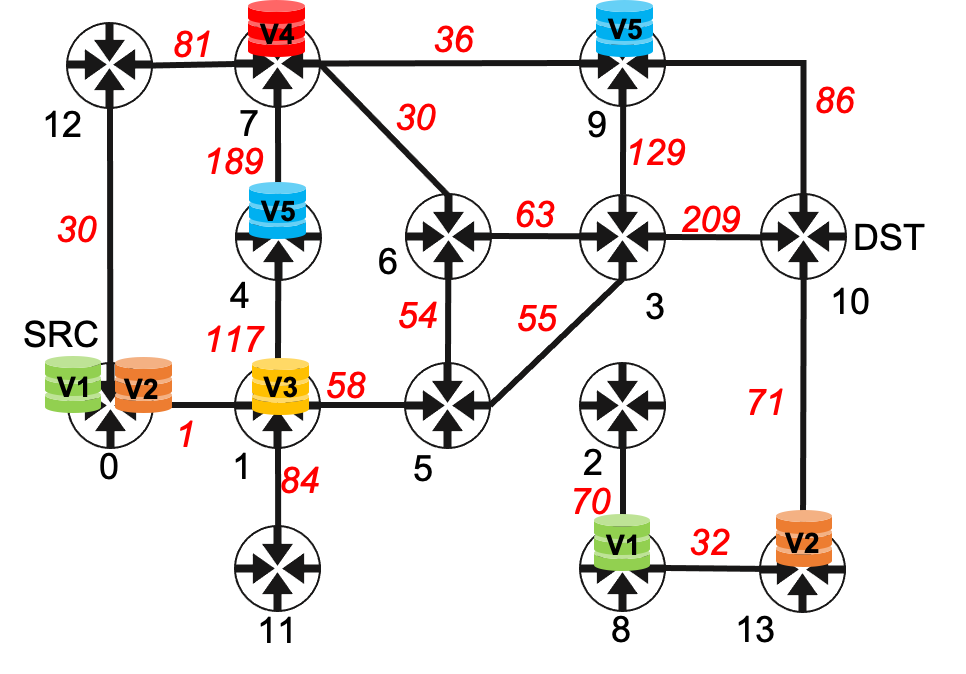}
    }
    \hbox{\small \hspace{0.05in}(b) Changed Topology Type(1)
    \hspace{0.13in} (c) Changed Topology Type(2)}
    \caption{Examples of changing topologies. (a) the original topology `internet2'. (b) and (c) show examples of changed topologies with two changing strategies, respectively. Black colored numbers on each node are indexes of the nodes. Red colored numbers on each edge are the delay times of the edges. Each colored box labeled as `V$+$number' refers to several types of deployed VNF instances.}
    \label{fig:topologies}
\end{figure}

\begin{table*}[t]
\caption{Experiment results of SL and RL learning approaches on 3 test cases}
\begin{center}
\begin{tabular}{|l|c|c|c|c|}
\hline
\textbf{Approach}&\multicolumn{2}{|c|}{
\textbf{\textit{Original Topo. Test}}}
&\textbf{\textit{Random Topo. Test}}
&\textbf{\textit{Random Topo.+VNFs Test}} \\
\cline{2-5} 
\textbf{} 
& \textbf{\textit{Failure Ratio}}
& \textbf{\textit{Delay Ratio}}
& \textbf{\textit{Failure Ratio (Deter. Rate)}}
& \textbf{\textit{Failure Ratio (Deter. Rate)}} \\
\hline
SL(Baseline) & 0.0080 & 1.1515 & 0.5133 (64.1) & 0.7399 (92.5)\\
\hline
RL($\lambda=0$) & \textbf{0.0064} & 1.4615 & 0.2627 (41.0) & 0.4410 (68.9) \\
\hline
RL($\lambda=1$) & 0.0072 & 1.4447 & 0.3008 (41.7) & 0.4753 (66.0) \\
\hline
RL($\lambda=0$) + CS1 & 0.0092 & 1.6150 & 0.0432 (4.7) & 0.0850 (9.2) \\
\hline
RL($\lambda=1$) + CS1 & 0.0105 & 1.4497 & 0.0420 (4.0) & 0.0941 (8.9) \\
\hline
RL($\lambda=0$) + CS2 & 0.0103 & 1.5823 & \textbf{0.0403} (3.9) & \textbf{0.0663} (6.4) \\
\hline
RL($\lambda=1$) + CS2 & 0.0135 & 1.3424 & 0.0496 (3.7) & 0.0817 (6.1) \\
\hline
\end{tabular}
\label{table:result_table}
\end{center}
\end{table*}

\subsection{Training on Random Topologies}
\label{subsec:random_topologies}
The SL approach \cite{Heo:supervised_gnn_sfc} stated that the reason for performance deterioration in the changed topology tests was because the model could not be trained on various topology where labels were not available. In our proposed RL method, we change the original topology randomly at each episode during training. For the experiments, we applied two strategies for changing the topology. 

First, we add a random number of nodes and edges at random locations. The probability and the number of trials to add a node and an edge are set to (0.1 and 12) and (0.3 and 15), respectively. Whenever we add a new node, for the node we added two edges with its neighbor nodes. After the adding processes, we remove randomly selected edges. The probability and the number of trials for removing edges are set to (0.3 and 30). When removing a selected edge disconnects any node, we dismiss the removal trial. Another strategy makes additional changes 
after applying the first strategy. We change the locations of the existing VNF instances randomly. Fig. \ref{fig:topologies} shows examples of changed topology using these strategies. 

With one of two strategies, we generated 100 random topologies before training and selected one structure randomly at the beginning of each episode for training. For testing, we also generated 100 random topologies using one of the strategies.

\section{Experiments and Results}
\label{sec:experiment_results}
\subsection{Experimental Settings}
We trained the GG-RNN model from scratch by SL  \cite{Heo:supervised_gnn_sfc} and used it as the pre-trained model for RL methods. 
The hyperparameters are the same as \cite{Heo:supervised_gnn_sfc} except for the learning rate of the proposed RL approaches, which is 0.00001. We used the $\epsilon$-greedy sampling for exploration where $\epsilon$ was 0.01. The discount factor $\gamma$ in Eq. \eqref{eq:returns} was set to 0.999. In the experiments, the models were trained and tested on a simulated dataset \cite{Heo:supervised_gnn_sfc}, which was generated in the same scenarios as \cite{lange2019predicting}.

We conducted three types of tests, (1) `Original Topo. Test', (2) `Random Topo. Test', and (3) `Random Topo.+VNFs Test': (1) the model generates paths on the topology `internet2' which we also used for training, (2) the model generates paths on randomly changed topology using the first strategy mentioned in section \ref{subsec:random_topologies},  and (3) the model generates on the same topologies as the second test, but the positions of the VNF instances are also changed randomly. 

The `Failure Ratio' is the ratio of the failed SFC requests out of all the SFC requests. The `Delay Ratio' is the total delay time of the generated path over the total delay time of the label path. Note that this metric is computable only for the first test, because the labels are not available for the other tests. The deterioration rate `Deter. Rate' is the ratio of `Failure Ratio' of the random tests (the second or third test) to `Failure Ratio' of the first test indicating how much the quality deteriorated because of topology change.

\subsection{Results and Analysis}
Table \ref{table:result_table} summarizes the experiment results comparing the proposed methods to the baseline method. In the first test (`Original Topo. Test'), our proposed methods have similar performance in terms of `Failure Ratio', though they are slightly lower for `Delay Ratio'. It is mainly because the reward for the RL methods is based on whether the model generates a valid path or not. Actually, `RL($\lambda$=0)' outperforms the baseline model for `Failure Ratio' since it was trained to minimize failures on the same topology.

We compared the RL methods trained with different topology change strategies. In Table \ref{table:result_table}, `CS1' and `CS2' indicate that the model was trained using topology change strategies 1 and 2, respectively, as in section \ref{subsec:random_topologies}. Although the increased complexity of the task lowers the quality of the models in the first test, they significantly outperform the baseline model in the second and the third tests. Moreover, the models trained by `CS2' demonstrated greater robustness than the models trained by `CS1' in the third test. These results confirm that the model trained on various topologies becomes more flexible and robust for different topologies.

To verify the effect of the delay penalty term in Eq. \eqref{eq:delay}, we compared the RL methods with different $\lambda$ values, 0 or 1. When $\lambda=1$, the reward includes delay time as a penalty and it decreases `Delay Ratio' compared to the models with $\lambda=0$, while `Failure Ratio' slightly increased. Simply stated, this is because the models were trained to maximize the reward, and this demonstrates that we can adapt $\lambda$ depending on various context of network management.

\section{Conclusion}
\label{sec:conclusion}
In this paper, to make the SFC module more flexible and robust for various network topologies, we proposed the RL methods to train the GNN based SFC model. 
The proposed methods could train the model on various topologies where the SL method could not because the labels are not available.
The experiment results demonstrated that the proposed approaches make the SFC module work on random topology tests where the baseline method does not work. 

In future, we can evaluate the methods in real computing network environments where a lot of different topologies exist without labels.

\vfill\pagebreak

\bibliographystyle{IEEEbib}
\bibliography{my_bib}

\begin{thebibliography}{10}

\bibitem{kreutz:sdn-survey}
Diego Kreutz, Fernando M.~V. Ramos, Paulo~Esteves Verissimo, Christian~Esteve
  Rothenberg, Siamak Azodolmolky, and Steve Uhlig,
\newblock ``Software-defined networking: A comprehensive survey,''
\newblock {\em Proceedings of the IEEE}, vol. 103, no. 1, pp. 14--76, Jan 2015.

\bibitem{mijumbi:nfv-survey}
Rashid Mijumbi, Joan Serrat, Juan-Luis Gorricho, Niels Bouten, Filip~De Turck,
  and Raouf Boutaba,
\newblock ``Network function virtualization: State-of-the-art and research
  challenges,''
\newblock {\em IEEE Communications Surveys Tutorials}, vol. 18, no. 1, pp.
  236--262, Firstquater 2016.

\bibitem{datta:intelligent-networking}
Avishek Datta, Aashi Rastogi, Oindrila~Ray Barman, Reynold D’Mello, and Omar
  Abuzaghleh,
\newblock ``An approach for implementation of artificial intelligence in
  automatic network management and analysis,''
\newblock {\em Lecture Notes in Networks and Systems}, vol. 22, pp. 901--909,
  Jan 2018.

\bibitem{Bhamare:sfc-survey}
Deval Bhamare, Raj Jain, Mohammed Samaka, and Aiman Erbad,
\newblock ``A survey on service function chaining,''
\newblock {\em J. Netw. Comput. Appl.}, vol. 75, no. C, pp. 138–155, Nov.
  2016.

\bibitem{pei:deep-sfc}
Jianing Pei, Peilin Hong, and Defang Li,
\newblock ``Virtual network function selection and chaining based on deep
  learning in sdn and nfv-enabled networks,''
\newblock in {\em 2018 IEEE International Conference on Communications
  Workshops (ICC Workshops)}, 2018, pp. 1--6.

\bibitem{li:ggnn}
Yujia Li, Daniel Tarlow, Marc Brockschmidt, and Richard Zemel,
\newblock ``Gated graph sequence neural networks,''
\newblock {\em arXiv}, 2015.

\bibitem{Heo:supervised_gnn_sfc}
DongNyeong Heo, Stanislav Lange, Hee-Gon Kim, and Heeyoul Choi,
\newblock ``Graph neural network based service function chaining for automatic
  network control,''
\newblock {\em ArXiv}, vol. abs/2009.05240, 2020.

\bibitem{scarselli:gnn}
Franco Scarselli, Marco Gori, Ah~Chung Tsoi, Markus Hagenbuchner, and Gabriele
  Monfardini,
\newblock ``The graph neural network model,''
\newblock {\em IEEE Transactions on Neural Networks}, vol. 20, no. 1, pp.
  61--80, 2009.

\bibitem{Williams:REINFORCE}
Ronald~J. Williams,
\newblock ``Simple statistical gradient-following algorithms for connectionist
  reinforcement learning,''
\newblock {\em Mach. Learn.}, vol. 8, no. 3–4, pp. 229–256, May 1992.

\bibitem{Ranzato2015}
Marc'Aurelio Ranzato, Sumit Chopra, Michael Auli, and Wojciech Zaremba,
\newblock ``{Sequence Level Training with Recurrent Neural Networks},''
\newblock {\em arXiv}, pp. 1--15, 2015.

\bibitem{lange2019predicting}
Stanislav Lange, Hee-Gon Kim, Se-Yeon Jeong, Heeyoul Choi, Jae-Hyung Yoo, and
  James Won-Ki Hong,
\newblock ``Predicting vnf deployment decisions under dynamically changing
  network conditions,''
\newblock in {\em 15th International Conference on Network and Service
  Management (CNSM)}, 2019, pp. 1--9.

\bibitem{cho:gru}
Kyunghyun Cho, Bart van Merrienboer, Caglar Gulcehre, Dzmitry Bahdanau, Fethi
  Bougares, Holger Schwenk, and Yoshua Bengio,
\newblock ``Learning phrase representations using rnn encoder-decoder for
  statistical machine translation,''
\newblock {\em arXiv}, 2014.

\end{thebibliography}

\end{document}